\documentclass[a4paper]{cas-sc}
\usepackage[numbers]{natbib}
\usepackage{multirow}
\usepackage{amsmath}
\usepackage{amsfonts}
\usepackage{bm}
\usepackage[noend]{algpseudocode}
\usepackage{algorithmicx,algorithm}
\usepackage{booktabs}
\usepackage{graphicx}
\usepackage{subfig}
\usepackage[misc]{ifsym}
\usepackage{epstopdf}
\usepackage{caption}
\usepackage{appendix}
\usepackage{placeins}
\usepackage{float}
\usepackage{emptypage}
\def\tsc#1{\csdef{#1}{\textsc{\lowercase{#1}}\xspace}}
\tsc{WGM}
\tsc{QE}
\tsc{EP}
\tsc{PMS}
\tsc{BEC}
\tsc{DE}

\newcounter{daggerfootnote}

\begin{document}
\begin{sloppypar}

\let\WriteBookmarks\relax
\def\floatpagepagefraction{1}
\def\textpagefraction{.001}
\let\printorcid\relax
\shortauthors{Yuanhao Chen et~al.}
\shorttitle{}
\renewcommand{\figurename}{Fig.}

\title [mode = title]{Sample-Efficient Policy Constraint Offline Deep Reinforcement Learning based on Sample Filtering}                      

\author[1]{Yuanhao Chen}
\author[2]{Qi Liu}
\corref{correspond}
\ead{liuqi@mail.neu.edu.cn}
\author[1]{Pengbin Chen}
\author[3]{Zhongjian Qiao}
\author[1]{Yanjie Li}
\corref{correspond}
\ead{ autolyj@hit.edu.cn}
\address[1]{Guangdong Key Laboratory of Intelligent Morphing Mechanisms and Adaptive Robotics and School of Inteligence Science and Engineering, the Harbin Institute of Technology Shenzhen, Shenzhen, 518055, China.}
\address[2]{Faculty of Robot Science and Engineering, Northeastern University, Shenyang, 110819, China.}
\address[3]{School of Systems Engineering,
City University of Hong Kong, Hong Kong, China.}

\cortext[correspond]{Corresponding author}

\begin{abstract}
Offline reinforcement learning (RL) aims to learn a policy that maximizes the expected return using a given static dataset of transitions. However, offline RL faces the distribution shift problem. The policy constraint offline RL method is proposed to solve the distribution shift problem. During the policy constraint offline RL training, it is important to ensure the difference between the learned policy $\pi$ and behavior policy $\pi_\beta$ within a given threshold. Thus, the learned policy $\pi$ heavily relies on the quality of the behavior policy $\pi_\beta$. However, a problem exists in existing policy constraint methods: if the dataset contains many low-reward transitions, the learned $\pi$ will be contained with a suboptimal reference policy $\pi_\beta$, leading to slow learning speed, low sample efficiency, and inferior performances. This paper shows that the sampling method in policy constraint offline RL that uses all the transitions in the dataset can be improved. A simple but efficient sample filtering method is proposed to improve the sample efficiency and the final performance. First, we evaluate the score of the transitions by average reward and average discounted reward of episodes in the dataset and extract the transition samples of high scores. Second, the high-score transition samples are used to train the offline RL algorithms. We verify the proposed method in a series of offline RL algorithms and benchmark tasks. Experimental results show that the proposed method outperforms baselines.

\end{abstract}


\maketitle

\section{Introduction}
\label{Introduction}
\begin{figure}[htbp]
	\centerline{\includegraphics[width=8cm, height=6cm]{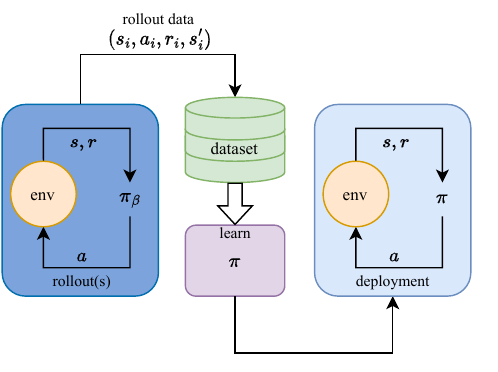}}
	\caption{Offline RL.}
	\label{Figure: Offline Reinforcement Learning}
\end{figure}

Deep reinforcement learning (RL) has achieved impressive successes in various challenging domains, such as robotic control \cite{liu2025sample,WANG2024106472,10508809}, multi-agent systems \cite{9812341,HUANG2024106547}, and video games \cite{liu2025corlhf,QI2023489,10466624}. Successful algorithms such as Deep Q-Networks \cite{mnih2015human}, Deep Deterministic Policy Gradient \cite{lillicrap2015continuous}, and Twin Delayed Deep Deterministic policy gradient algorithm (TD3) \cite{fujimoto2018addressing} are applied to tackle intricate robotic control problems. However, the above online deep RL algorithms require the agent to interact with the environment to collect transition samples during policy learning, slowing the learning process and leading to low sample efficiency. Moreover, the collection of transition samples may be challenging, hazardous, and costly in various industries, such as healthcare, finance, autonomous driving, and robotics \cite{prudencio2023survey}. To solve the above problems of online RL algorithms, the data-driven offline RL \cite{levine2020offline} method is proposed. Offline RL aims to learn a policy that maximizes the expected return using a given static dataset of transitions. In offline RL, the transition samples in the static dataset are obtained from the interactions between multi-unknown behavior policies and the environment. As shown in Fig. \ref{Figure: Offline Reinforcement Learning}, in offline RL learning, the transition samples in the dataset are used for training offline RL methods \cite{levine2020offline,prudencio2023survey}. Because there is no need to interact with the environment synchronously during learning, the sample efficiency is improved, and the learning speed is accelerated. 

However, research on offline RL faces numerous challenges. One of the core challenges is the distribution shift problem \cite{levine2020offline} because the distribution of the provided transitions may not match the distribution of states in the test task. This is different from online deep RL algorithms, where the algorithms interact with the environment, enabling them to explore and discover optimal solutions while refining the estimation of the $Q$-function under the current policy. This mitigates the risk of incorrect value estimation for unfamiliar state-action pairs by the $Q$-function. Thus, for offline RL, due to the lack of exploration in an offline dataset, the learned policy, value function, and model might be trained under a specific distribution, subsequently being evaluated under a distinct distribution. The learned $Q$-function fails to accurately evaluate the value of actions within alternative distributions, leading to erroneous reward estimations. Thus, the trained policy becomes unable to select optimal actions, potentially converging towards suboptimal or even detrimental outcomes rather than achieving the desired optimal solution. This can be intuitively understood as the unlearning of the new environment, and therefore, it has no reason to provide the correct answers. 

Existing research in offline RL primarily addresses the distribution shift problem; These methods can be classified into three categories: policy constraint methods \cite{fujimoto2019off}, value regularization methods \cite{kumar2020conservative}, and uncertainty-based methods \cite{kumar2019stabilizing}. This paper focuses on policy constraint methods that are widely studied and utilized \cite{fujimoto2019off}. Fujimoto et al. \cite{fujimoto2019off} proposed the policy constraint offline RL method to solve the distribution shift problem. After that, numerous policy constraint offline RL variants \cite{kumar2019stabilizing,wu2019behavior,fujimoto2021minimalist} have been proposed. In these methods, the core insight is to ensure a small difference between the learned policy $\pi$ and the reference (behavior) policy 
 $\pi_\beta$. 

However, existing policy constraint methods face a noteworthy challenge: Policy constraint offline RL methods constrain the learned policy $\pi$ entirely based on the constraint $D(\pi,\pi_\beta) \leq \epsilon$ over all the data in the dataset. If the reference policy $\pi_{\beta}$ collects numerous low-reward samples, the learned policy $\pi$ will also learn from these inferior samples. This constraint leads to limiting final performances, slow learning speed, and low sample efficiency. In this paper, we improve the policy constraint problems in policy constraint offline RL algorithms.

The main contributions of this work can be summarized as follows:
\begin{itemize}
	\item In this paper, we show that the sampling method in policy constraint offline RL that uses all the transition data in the dataset can be improved. A simple yet efficient sample filtering method is proposed to improve the sample efficiency and the final performance. Specifically, first, we evaluate the score of the transitions in the dataset and extract the transition samples of high scores. Second, the high-score transition samples are used to train the offline RL algorithms.
	\item We verify the proposed method in a series of offline RL algorithms and benchmark tasks. Experimental results show that the proposed method outperforms baselines.
\end{itemize}
\section{Related Work}
\label{Section: Related work}

Within the realm of offline reinforcement learning, two branches exist: model-free offline RL and model-based offline RL \cite{kidambi2020morel}. This paper focuses on model-free offline RL due to its extensive research and superior performance compared to model-based offline RL. The model-free offline RL can be categorized into three classifications: policy constraint methods \cite{fujimoto2019off}, value regularization methods \cite{kumar2020conservative}, and uncertainty-based methods \cite{kumar2019stabilizing}.

\subsection{Policy Constraint Methods}
This approach maintains proximity between learned $\pi$ and behavior policy $\pi_{\beta}$ to ensure accurate $Q$-value estimation \cite{fujimoto2019off}, formally constrained as $D(\pi, \pi_{\beta}) \leq \epsilon$ where $D$ is a divergence metric (e.g., $f$-divergence). Batch-Constrained deep Q-learning (BCQ) \cite{fujimoto2019off} employs a VAE \cite{kingma2013auto} to generate dataset-like actions, selecting maximum-$Q$ actions from perturbed outputs to ensure action similarity. Building on BCQ, Bootstrapping error accumulation reduction (BEAR) \cite{10.5555/3454287.3455342} replaces perturbation with a support constraint using Maximum Mean Discrepancy (MMD) \cite{10.5555/2503308.2188410} between policy-sampled actions and those from an approximated behavior policy, achieving greater distributional flexibility. Twin delayed deep deterministic policy gradient with behavior cloning (TD3+BC) \cite{fujimoto2021minimalist} simplifies constraint via direct behavior cloning loss on policy outputs, anchoring actions to dataset samples without generative components. Diverging from explicit constraints, Implicit Q-Learning (IQL) \cite{kostrikov2021offline} implements implicit regularization through conservative $Q$-learning: it trains $Q$-functions to upper-bound dataset action values via expectile regression, then extracts policies by maximizing these conservative estimates. These constraint-based methods significantly enhance offline RL stability and performance \cite{kumar2019stabilizing, wu2019behavior}.

\subsection{Value Regularization Methods}
Similar to policy constraint methods, the distance $D(\pi, \pi_{\beta})$ in value regularization can be treated as a penalty term added to the objective or reward functions. Regularization is an effective technique that permits fine-tuning of the learned function by introducing a penalty. By employing value regularization, the method imposes a penalty on the learned value function, thereby encouraging more conservative estimates. \cite{kumar2020conservative} added a regularizer on the top of the value function, learning a conservative Q-function, which is able to yield a true lower bound on the value under the current policy, enabling both policy evaluation and policy improvement processes. \cite{yu2021combo} provided a model-based regularization method, which can learn a dynamics model by interpolating model-free and model-based components. However, regularization terms are usually less strict compared to policy constraints since they do not impose restrictions on the policy $\pi_{\beta}$. To avoid out-of-distribution (OOD) actions, value regularization methods are commonly supplemented with other techniques like conservative models or policy constraints. These additional methods effectively deter the selection of actions outside the desired distribution.

\subsection{Uncertainty-based Methods}
Uncertainty-based offline RL methods \cite{kumar2019stabilizing} provide flexibility in switching between naive and conservative off-policy RL methods, depending on the level of trust in the models' generalization ability. \cite{NEURIPS2020_a322852c} introduces a penalty term in the uncertainty estimation of the model, reducing reliance on inaccurate model predictions. \cite{NEURIPS2021_3d3d286a} took into account the confidence in Q-value prediction, proposing an uncertainty-based offline reinforcement learning. These methods entail estimating the uncertainty of the approximations, such as the model, policy, or value function. By doing so, the constraints on the learned policy in regions with low uncertainty can be loosened. Uncertainty estimation methods focus on defining the distribution and the corresponding uncertainty estimator, which are necessary for evaluating the objective.

\section{Preliminaries}
\label{Section: Preliminary}
This paper considers the RL modeled as a Markov Decision Process (MDP), modeled by a tuple $(\mathcal{S}, \mathcal{A}, \mathcal{P}, \mathcal{R}, \gamma, T)$ \cite{sutton2018reinforcement}. $\mathcal{S}$ represents the state space, $\mathcal{A}$ denotes the action space, $\mathcal{P}: \mathcal{S} \times \mathcal{A} \times \mathcal{S} \rightarrow [0,1]$ represents the state transition probability. $\mathcal{R}: \mathcal{S} \times \mathcal{A} \rightarrow R$ represents a reward function, $\gamma \in [0,1)$ represents a discount factor, and $T$ is the time horizon. At each timestep $t$, an action $a_{t} \in \mathcal{A}$ is determined by a policy. After transiting into the next state by sampling from $p\left(s_{t+1}, r(s_{t}, a_{t}) \mid s_{t}, a_{t}\right)$, where $p \in \mathcal{P}$, the agent obtains a scalar reward $r\left(s_{t}, a_{t}\right) \in \mathcal{R}$. The agent keeps interacting with the environment until it enters a terminal state. RL aims to learn a policy $\pi: \mathcal{S} \times \mathcal{A} \rightarrow [0, 1] $ for decision-making problems by maximizing discounted cumulative rewards. For a policy $\pi$, the optimized objective function $J(\pi)$ is
\begin{equation}
J(\pi)={\mathbb{E}^{\pi}}\left[\sum_{t=0}^{T} \gamma^{t} r\left(s_{t}, a_{t}\right)\right]
\label{Eq: objective function}
\end{equation}
where ${\mathbb{E}^{\pi}}$ denotes the expectation under policy $\pi$. The state-action value function ($Q$ function) is
\begin{equation}
Q^{\pi}(s, a)={\mathbb{E}^{\pi}}\left[\sum_{t=0}^{T} \gamma^{t} r\left(s_{t}, a_{t}\right) \mid S_t=s, A_t=a\right]
\label{Eq: state-action value function}
\end{equation}
RL methods update the $Q$ function using the recursive relationship in Bellman equation \cite{sutton2018reinforcement}:
\begin{equation}
Q^{\pi}\left(s_{t}, a_{t}\right)=\mathbb{E}^{\pi}\left[r\left(s_{t}, a_{t}\right)+\gamma \mathbb{E}_{a_{t+1} \sim \pi}\left[Q^{\pi}\left(s_{t+1}, a_{t+1}\right)\right]\right]
\label{Eq: recursive relationship in Bellman equation}
\end{equation}
\section{Sample-Efficient Policy Constraint Offline Deep RL based on Sample Filtering}
\label{Section: Sample-Efficient Policy Constraint Offline Deep Reinforcement Learning based on Sample Filtering}

\subsection{Motivation}
\label{Subsection：Motivation}

\subsubsection{Dataset Construction}
\label{Subsection：Dataset Construction}
In practical application scenarios, it is uncommon to have datasets that are exclusively generated by random, expert, or medium policies. In other words, most datasets are a mixture of random, expert, and medium policies. Thus, reconstructing a proper dataset for the experiment is crucial, as this choice plays a pivotal role in ensuring the realism and soundness of the resulting outcomes. This paper employed a hybrid approach, combining datasets from three policy types (random, expert, and medium) to construct a dataset that better reflects real-world scenarios. 

The method chooses three settings of dataset: random, expert, and medium \cite{fu2020d4rl}:
\begin{itemize}
    \item Random Dataset: a dataset sampled by a random behavior policy and serves as a naive baseline for dataset collection.
    \item Expert Dataset: a dataset collected with the final greedy expert policy by an online policy, which has been trained until convergence.
    \item Medium Dataset: a dataset between random and expert collected by a medium behavior policy that is stopped early while training.
\end{itemize}

\subsubsection{Defect of offline RL}
\label{Subsection：Defect of offline RL}

Offline RL policy is thoroughly trained from an existing dataset and cannot interact with the environment anymore. Existing offline RL scholars have proposed different methods to address the OOD problem. Offline RL methods incorporating policy constraints, such as BEAR and TD3+BC, can be defined using Eq. (\ref{eq 4}), which constrains the learned policy $\pi$ by enforcing $D(\pi,\pi_\beta) \leq \epsilon$. This will result in them being influenced by low-quality data.

\begin{equation}
\begin{array}{c}
\pi_{k+1} \leftarrow \underset{\pi}{\operatorname{argmax}} \mathbb{E}_{s \sim D}\left[\mathbb{E}_{a \sim \pi(a \mid s)}\left[\hat{Q}_{k+1}^{\pi}(s, a)\right]\right] \\
s.t. D\left(\pi,\pi_\beta\right)\leq\epsilon
\end{array}   
\label{eq 4}
\end{equation}

Other offline RL methods, like IQL, avoid querying out-of-sample actions. However, these methods remain susceptible to interference from poor-quality data within the dataset. This paper proposes a sample filtering method as a solution to mitigate this problem.

\subsection{Offline Deep RL with Sample Filtering Method}
\label{Subsection: Sample Filtering Method}

This paper proposes a sample filtering method on Offline RL that evaluates the original dataset, identifies and removes data with inferior performance, and selectively keeps those with superior performance. This process results in a refined dataset that is both superior in quality and reduced in size. The method can expedite the learning process and finally achieve improved results by using this filtered dataset.

\begin{figure}[htbp]
	\centerline{\includegraphics[width=10cm, height=5.4cm]{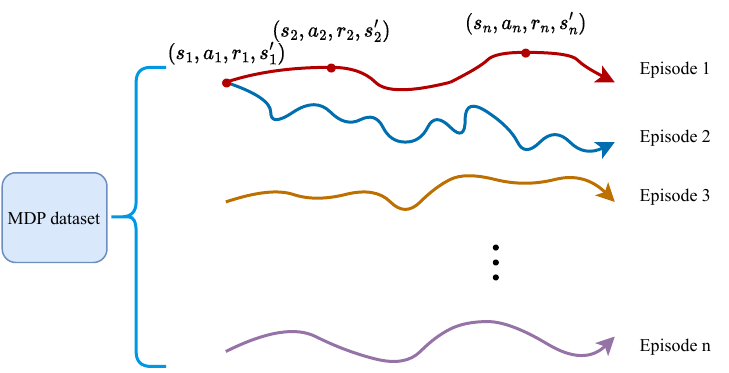}}
	\caption{Environmental episodes.}
	\label{Figure: Environmental Schematic Diagram}
\end{figure}

First, the smallest unit of data quality judgment should be clarified. In the case of an MDP dataset, the dataset is structured hierarchically into episodes and transitions, as depicted in Fig. \ref{Figure: Environmental Schematic Diagram}. Each MDP dataset consists of multiple episodes, with each episode including a set of several transitions. When determining the quality of a transition, a straightforward method is to estimate its reward directly. However, relying solely on individual transitions may lead to misjudgments, as the ultimate training outcome is determined by the sequence of transitions rather than individual ones. Hence, to accurately assess quality, we consider multiple transitions in their chronological order. This paper evaluates a transition by assessing the quality of the episode to which it belongs. In the evaluation process, if an episode is considered to be of high quality or low quality, this judgment extends to all transitions within that particular episode. This evaluation process is also consistent with reality as motions in reality are always in sequences.

\begin{algorithm}[htbp]
    \caption{Offline Deep RL with Sample Filtering Process}
    \textbf{Input}: \\
    $B$: the offline dataset
    \begin{algorithmic}[1] 
        \State Episode $b_i$ in dataset $B$ can be defined in chronological order as:
        $b_i=[(s_{1i},a_{1i},r_{1i},s'_{1i}),\cdots,(s_{ni},a_{ni},r_{ni},s'_{ni})]$
        \State If using $R_{avg}$ as judgement criteria, calculate the average reward of each episode $b_i$ as $R_{avgi}$, and consider episode $b_i$ superior while $R_{avgi}$ is higher than $R_{aver}$. The equation is shown in Eq. (\ref{R_avg}).
        \State If using $R_{disc}$ as judgement criteria, calculate the average discounted reward of each episode $b_i$ as $R_{disci}$, and consider episode $b_i$ superior while $R_{disci}$ is higher than $R_{disc}$. The equation is shown in Eq. (\ref{R_disc}).
        \State Get $B_{superior}$ and $B_{inferior}$ after measurement and the equation is showed in Eq. (\ref{bi_judgement}).
        \State Reassemble $B_{superior}$ into a new dataset $B'$ , and remove $B_{inferior}$.
        \State Apply dataset $B'$ to offline Deep RL process and obtain policy $\pi_F$
    \end{algorithmic}
    \textbf{Output}: $\pi_F$
    \label{alg: 1}
\end{algorithm}

The direct evaluation of a transition's quality can be achieved through the reward value assigned by the environment. Consequently, the method uses the single-episode-cumulative average reward or average discounted reward as a criterion to assess its quality, as average reward is a straight forward method to measure a distribution and discounted reward is widely used to measure a state in RL training. In this situation, no matter how long the episode is, we can still obtain its criterion. To mitigate the influence of low-value transitions, our method eliminates all transitions judged to be of inferior quality and reassembles the remaining transitions for training purposes. The average reward is defined as $R_{avg}$ and the average discounted reward is defined as $R_{disc}$, which is shown at Eq. (\ref{R_avg}) and Eq. (\ref{R_disc})

\begin{equation}
\begin{matrix}
R_{avg}=\frac{1}{k}\sum_{i=1}^{k}R_{avgi}
\\
s.t. R_{avgi}= \frac{1}{n}\sum_{j=1}^{n}r_{ji} 
\end{matrix}
\label{R_avg}
\end{equation}

\begin{equation}
\begin{matrix}
R_{disc}=\frac{1}{k}\sum_{i=1}^{k}R_{disci}
\\
s.t. R_{disci}=\frac{1}{n}\sum_{j=1}^{n}\sum_{h=j}^{n}\gamma^hr_{hi}
\end{matrix}
\label{R_disc}
\end{equation}

\noindent where $n$ is the number of transitions in an episode and $k$ is the number of episodes.

\begin{figure}[htbp]
	\centerline{\includegraphics[width=10cm, height=2.3cm]{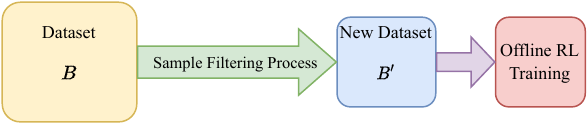}}
	\caption{Policy constraint offline deep RL based on sample filtering.}
	\label{Figure: Policy Constraint Offline Deep Reinforcement Learning based on Sample Filtering}
\end{figure}

The sampling method is shown in Eq. (\ref{bi_judgement}), where $B_{superior}$ is a set of superior episodes and $B_{inferior}$ is a set of inferior episodes.

\begin{equation}
\begin{matrix}
b_i\in \left\{\begin{matrix}
B_{superior} & &(R_{i}> R)
 \\
B_{inferior} & &(R_{i}\le R)
\end{matrix}\right.
 \\
s.t. R_{i}, R=R_{avgi}, R_{avg}\ or\ R_{disci}, R_{disc}
\end{matrix}
\label{bi_judgement}
\end{equation}
After applying the sample filtering process, we obtained a new dataset $B'$. Subsequently, we used this revised dataset $B'$ in the standard offline RL process. Fig. \ref{Figure: Policy Constraint Offline Deep Reinforcement Learning based on Sample Filtering} and Algorithm \ref{alg: 1} illustrate the entire process of offline deep RL with sample filtering.

\subsection{Discussions}
\label{Subsection: Discussions}
An intuitive puzzle of the proposed sample filtering method is that when the number of superior transition samples in the dataset is small, the number of inferior transition samples is large. In this case, there are always relatively superior and relatively inferior samples in the dataset. Therefore, the imbalance in quantities mentioned earlier does not occur. Thus, the proposed sample filtering method can also work well.

Another puzzle of the proposed method is why we should choose offline RL but not behavioral cloning (BC) \cite{kumar2022should}. BC is a supervised learning approach that learns the policy directly from expert state-action pairs, but offline RL is usually based on value function or policy-based methods that use pre-collected data to learn the optimal policy. BC imitates expert behavior so that similar actions are generated in a given state and it relies on expert demonstration data, which usually comes from one or multiple expert policies. In contrast, offline RL maximizes cumulative rewards by learning the optimal policy to perform efficiently with data from multiple policies, so that offline RL often has better generalization ability than BC. Additionally, as mentioned before, we can only obtain mixed datasets in reality which makes it inappropriate to use BC. Finally, it is proved that offline RL outperforms BC on noisy-expert datasets which is more likely to be in reality. 

This paper also tried to impose constraints on superior-quality data while not imposing constraints on inferior-quality data. However, this method did not yield satisfactory results. Further elaboration on this matter is beyond the scope of this discussion.

\begin{table}[htbp] 
    \centering
    \caption{Details of datasets}
    \label{tab: Details of Datasets}
    \begin{tabular}{c|c|c}
        \hline
        Environment           & Action Space                      & Observation Shape \\
        \hline
        Ant                & Box(-1.0, 1.0, (8,), float32)     & (27,)             \\
        HalfCheetah        & Box(-1.0, 1.0, (6,), float32)     & (17,)             \\
        Hopper             & Box(-1.0, 1.0, (3,), float32)     & (11,)             \\
        Walker2d           & Box(-1.0, 1.0, (6,), float32)     & (17,)             \\
        \hline
    \end{tabular}
\end{table}

\begin{figure}[htbp]
    \centering   
    \subfloat[Ant]{ 
    \begin{minipage}[b]{0.19\textwidth}
        \centering  
        \includegraphics[width=3cm, height=3cm]{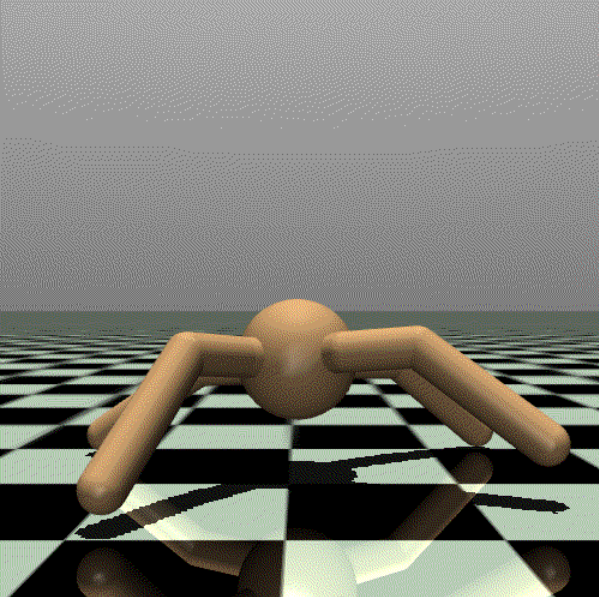} 
    \end{minipage}} 
    \subfloat[HalfCheetah]{ 
    \begin{minipage}[b]{0.19\textwidth}
        \centering  
        \includegraphics[width=3cm, height=3cm]{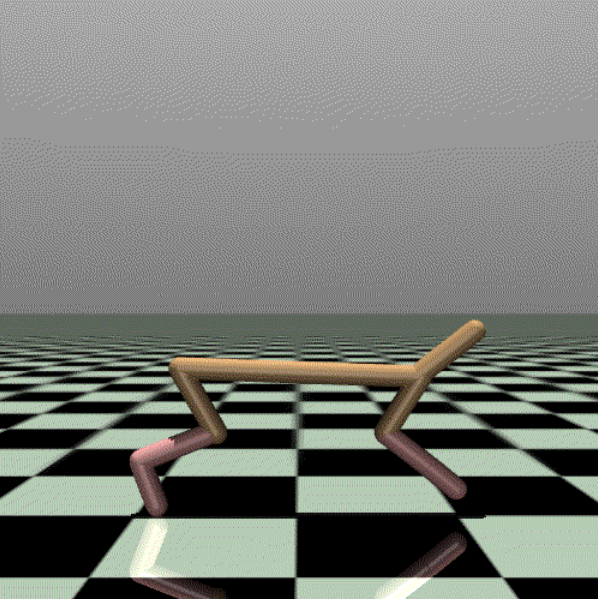} 
    \end{minipage}} 
    \subfloat[Hopper]{ 
    \begin{minipage}[b]{0.19\textwidth}
        \centering  
        \includegraphics[width=3cm, height=3cm]{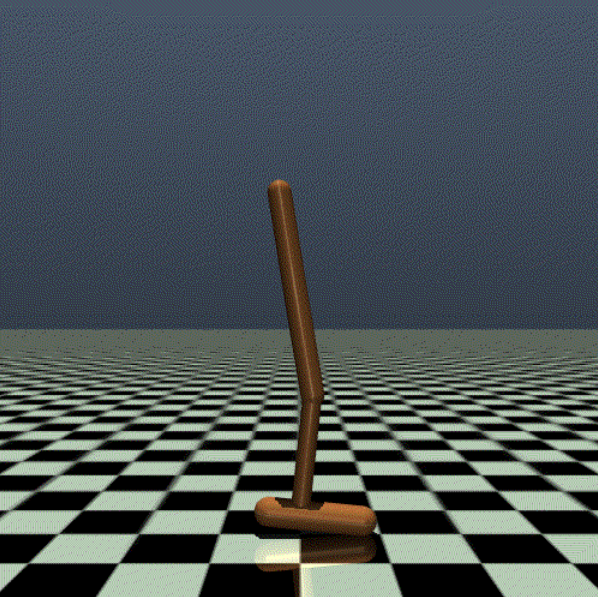} 
    \end{minipage}} 
    \subfloat[Walker2d]{ 
    \begin{minipage}[b]{0.19\textwidth}
        \centering  
        \includegraphics[width=3cm, height=3cm]{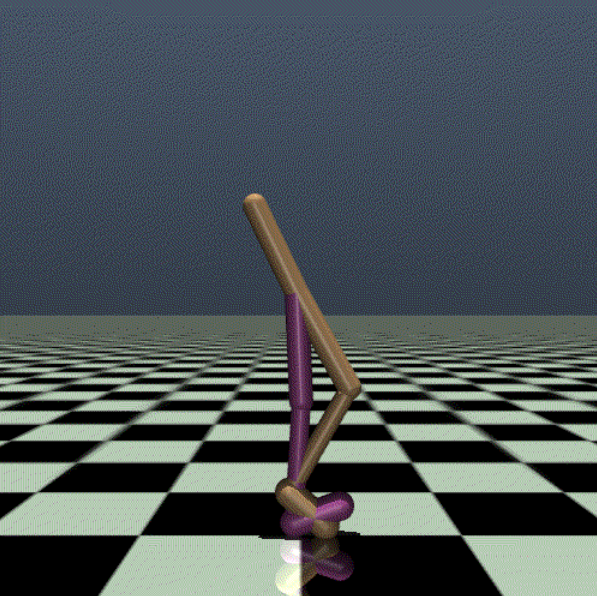} 
    \end{minipage}} 
    \caption{D4RL environment} 
    \label{Figure: D4RL environment} 
\end{figure}

\section{Experiments}
\label{Section: Experimental Results}

\subsection{Preparation}
\label{Section: Preparation}

\subsubsection{Dataset Selection}
\label{Section: Dataset Selection}

This experiment chose the D4RL\cite{fu2020d4rl} dataset, which is shown in Fig. \ref{Figure: D4RL environment}. Table \ref{tab: Details of Datasets} presents each environment's action and observation space details. The reason for choosing the D4RL dataset is that it has the following advantages:
\begin{itemize}
    \item D4RL is widely recognized as a benchmark dataset in the field of offline RL.
    \item It can provide multiple environments, so that there is no limit to verifying the quality of an algorithm, and multiple comparisons can be made to draw more reliable conclusions.
    \item It can provide data collected by different policies in the same environment.
\end{itemize}

\subsubsection{Algorithm Selection}
\label{Section: Algorithm Selection}

This experiment employed three algorithms to validate the proposed method: BEAR, TD3+BC, and IQL. These three methods are widely recognized in offline reinforcement learning for their comparatively good performance. Both BEAR and TD3+BC employ policy constraint technology to address the OOD problem. The TD3+BC algorithm extends the TD3 algorithm by incorporating behavioral cloning, enabling it to handle challenges in offline learning. On the other hand, BEAR mitigates the distribution shift by constraining policy updates within the support of the behavior policy, using MMD to enforce theoretical safety guarantees. As mentioned earlier, policy constraint methods can be perturbed by poorer-quality data in the dataset. Therefore, selecting these two methods allows us to evaluate the effectiveness of the proposed method. As for IQL, it does not utilize a policy constraint method to address OOD problems. However, considering that datasets commonly have a substantial impact on offline RL results, this experiment chose IQL to validate the method's effectiveness in the context of non-policy-constraint algorithms. 
\begin{figure}[htbp] 
    \centering
    \subfloat[BEAR-Ant]{ 
    \begin{minipage}[b]{0.5\textwidth}
        \centering  
        \includegraphics[width=6cm, height=4.2cm]{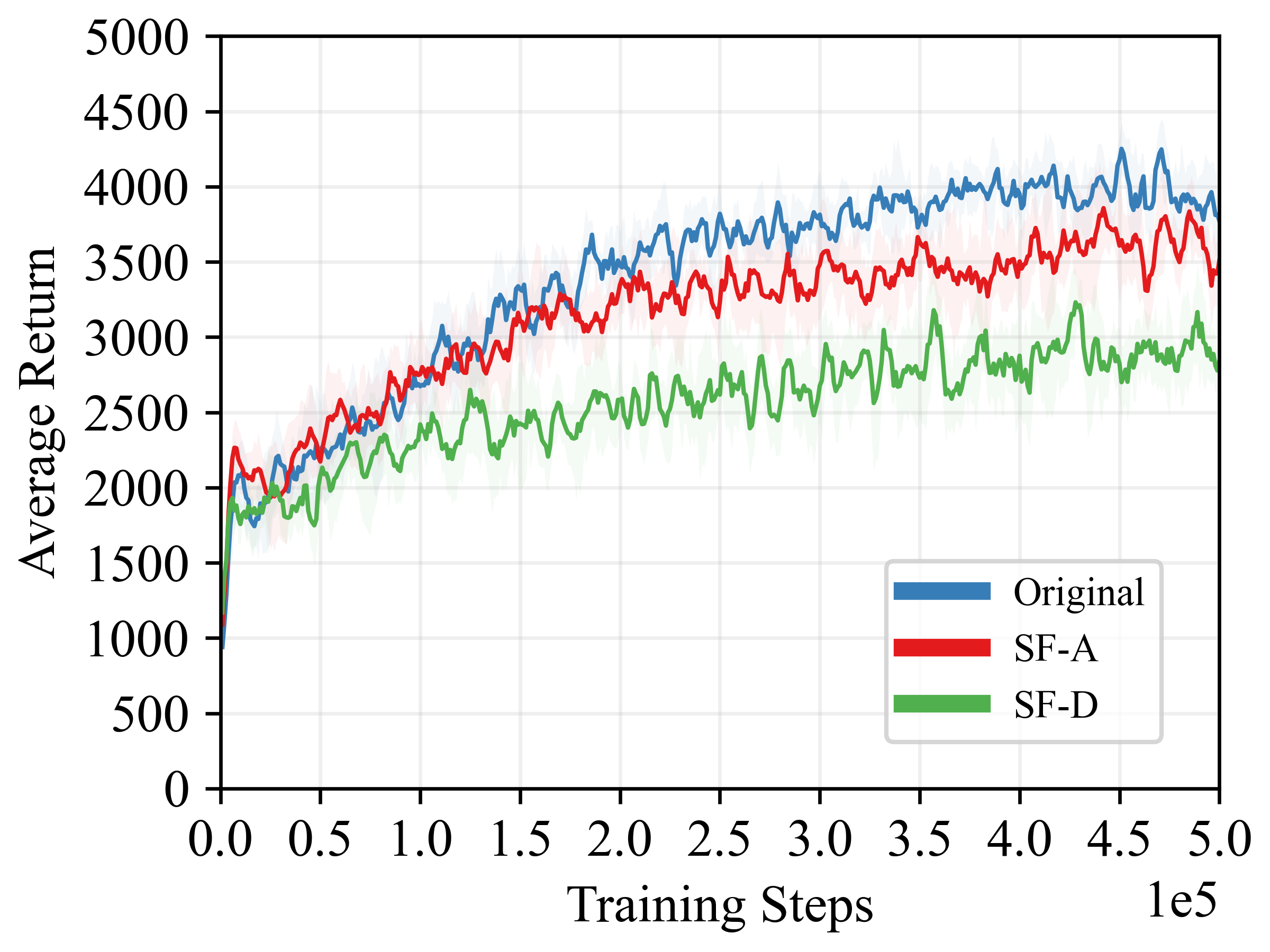} 
    \end{minipage}} 
    \subfloat[BEAR-HalfCheetah]{ 
    \begin{minipage}[b]{0.5\textwidth}
        \centering  
        \includegraphics[width=6cm, height=4.2cm]{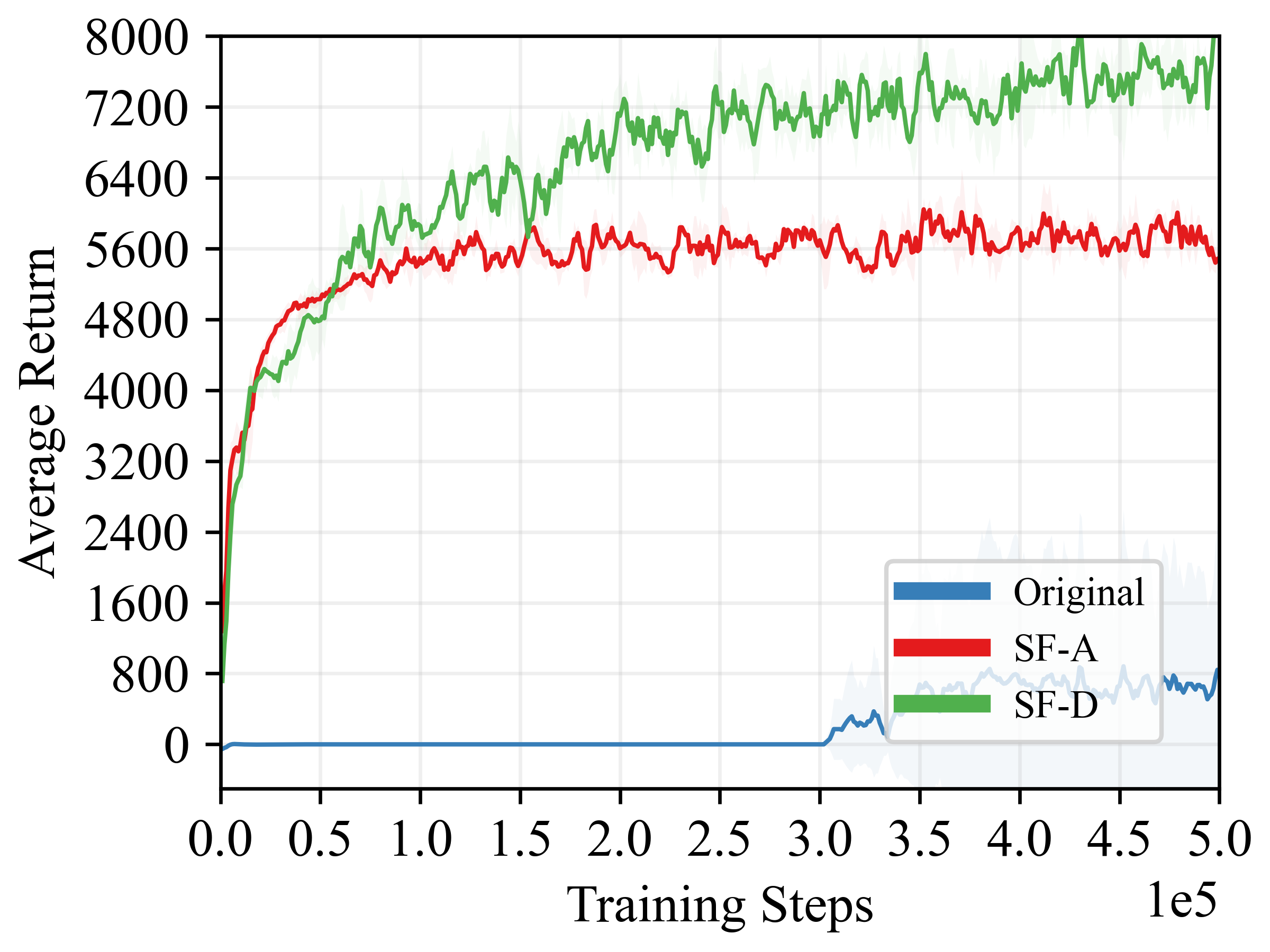} 
    \end{minipage}} 
    \\
    \subfloat[BEAR-Hopper]{ 
    \begin{minipage}[b]{0.5\textwidth}
        \centering  
        \includegraphics[width=6cm, height=4.2cm]{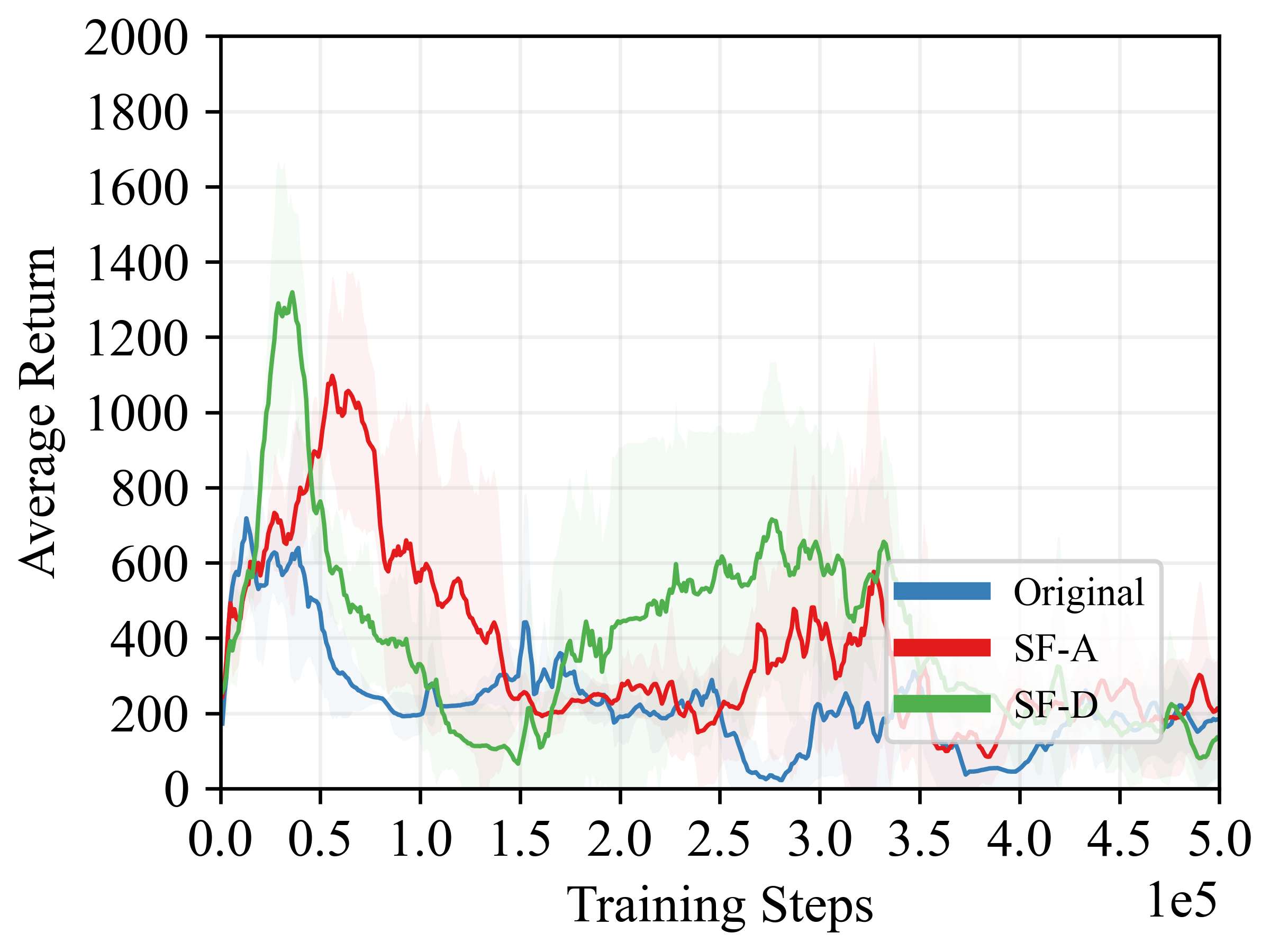} 
    \end{minipage}} 
       \subfloat[BEAR-Walker2d]{ 
    \begin{minipage}[b]{0.5\textwidth}
        \centering  
        \includegraphics[width=6cm, height=4.2cm]{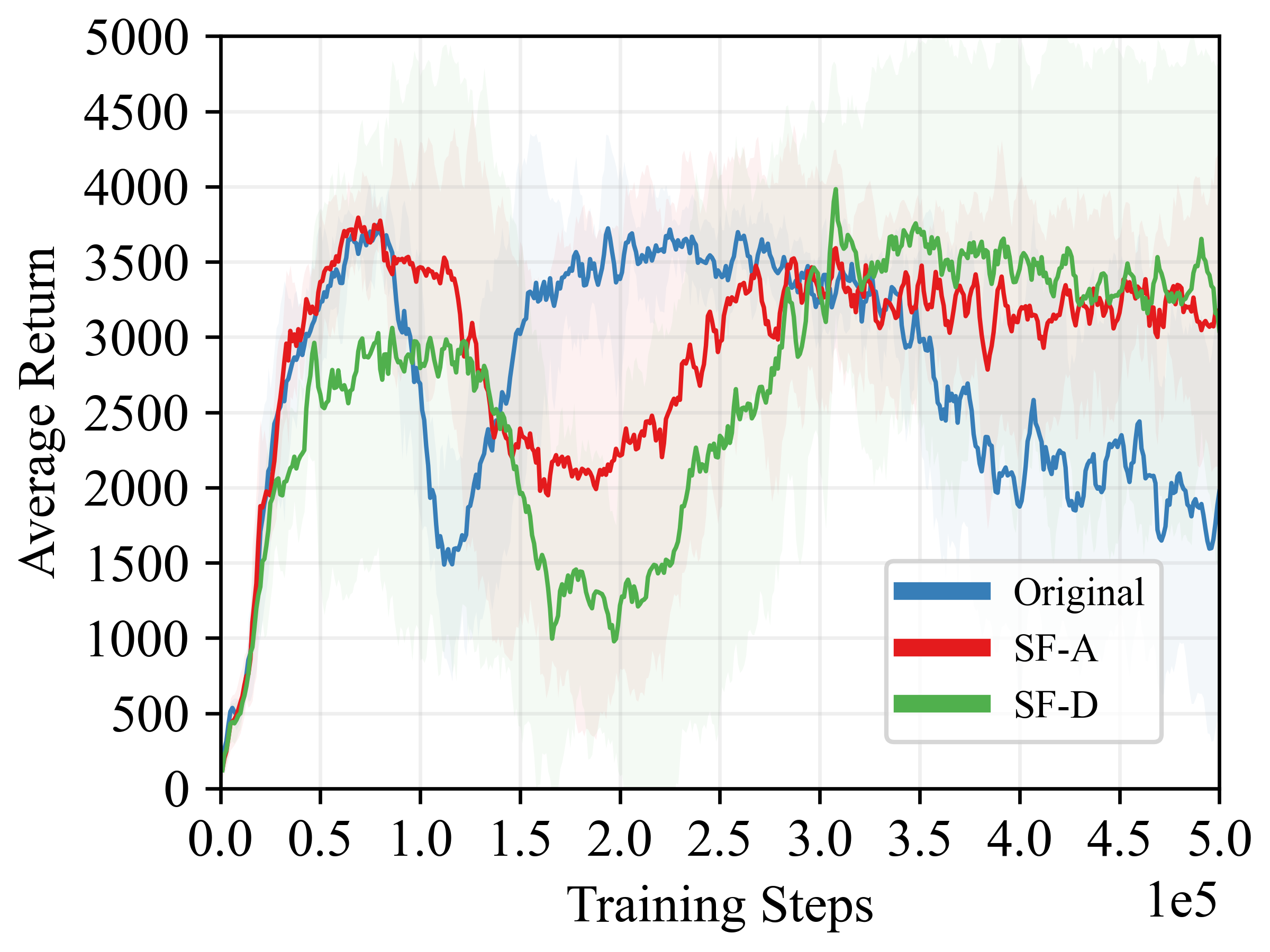} 
    \end{minipage}} 
    \caption{Experimental results of BEAR.} 
    \label{Figure: Result BEAR} 
\end{figure}

\begin{figure}[htbp] 
    \centering
    \subfloat[IQL-Ant]{ 
    \begin{minipage}[b]{0.5\textwidth}
        \centering  
        \includegraphics[width=6cm, height=4.2cm]{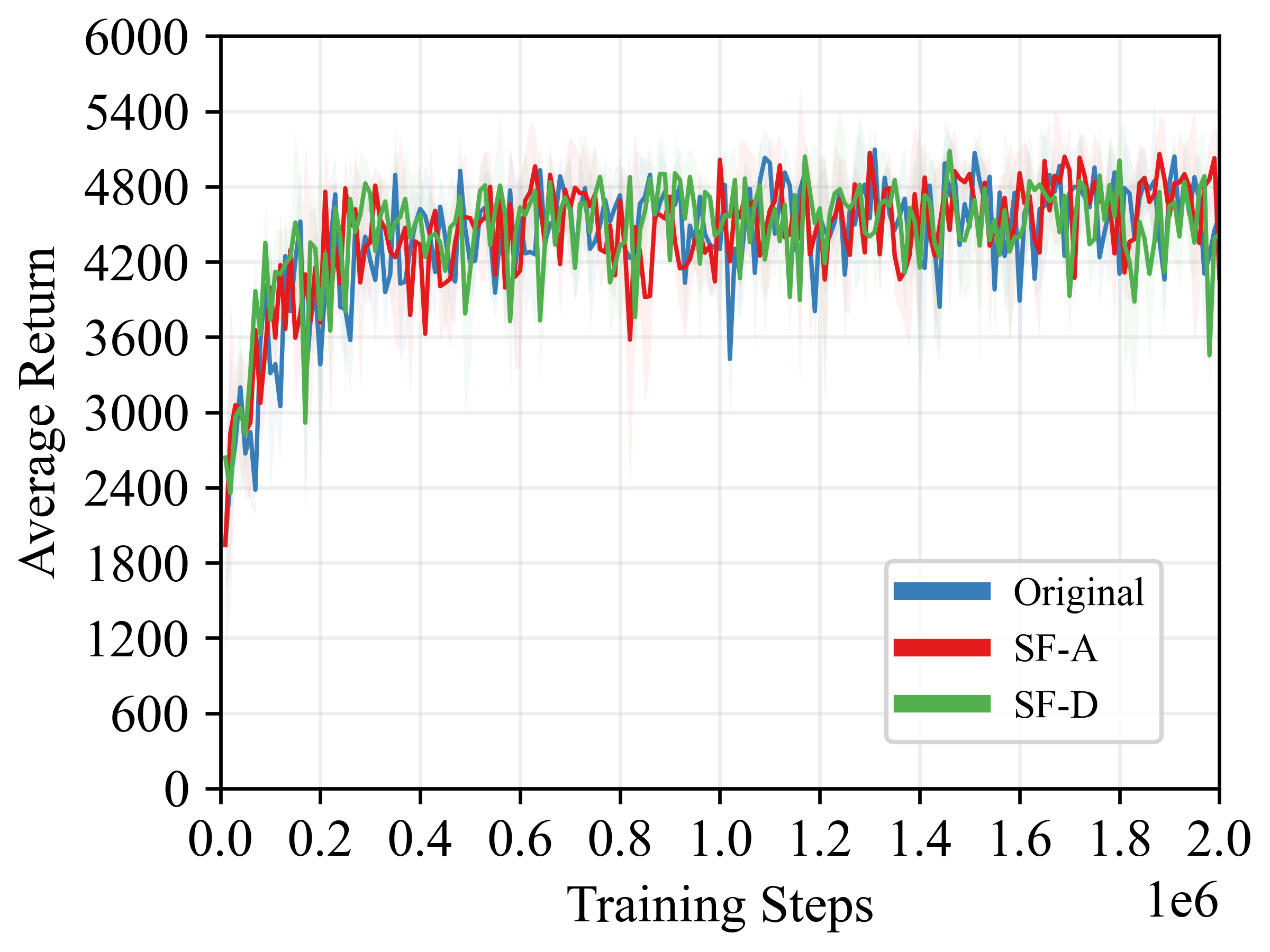} 
    \end{minipage}} 
    \subfloat[IQL-HalfCheetah]{ 
    \begin{minipage}[b]{0.5\textwidth}
        \centering  
        \includegraphics[width=6cm, height=4.2cm]{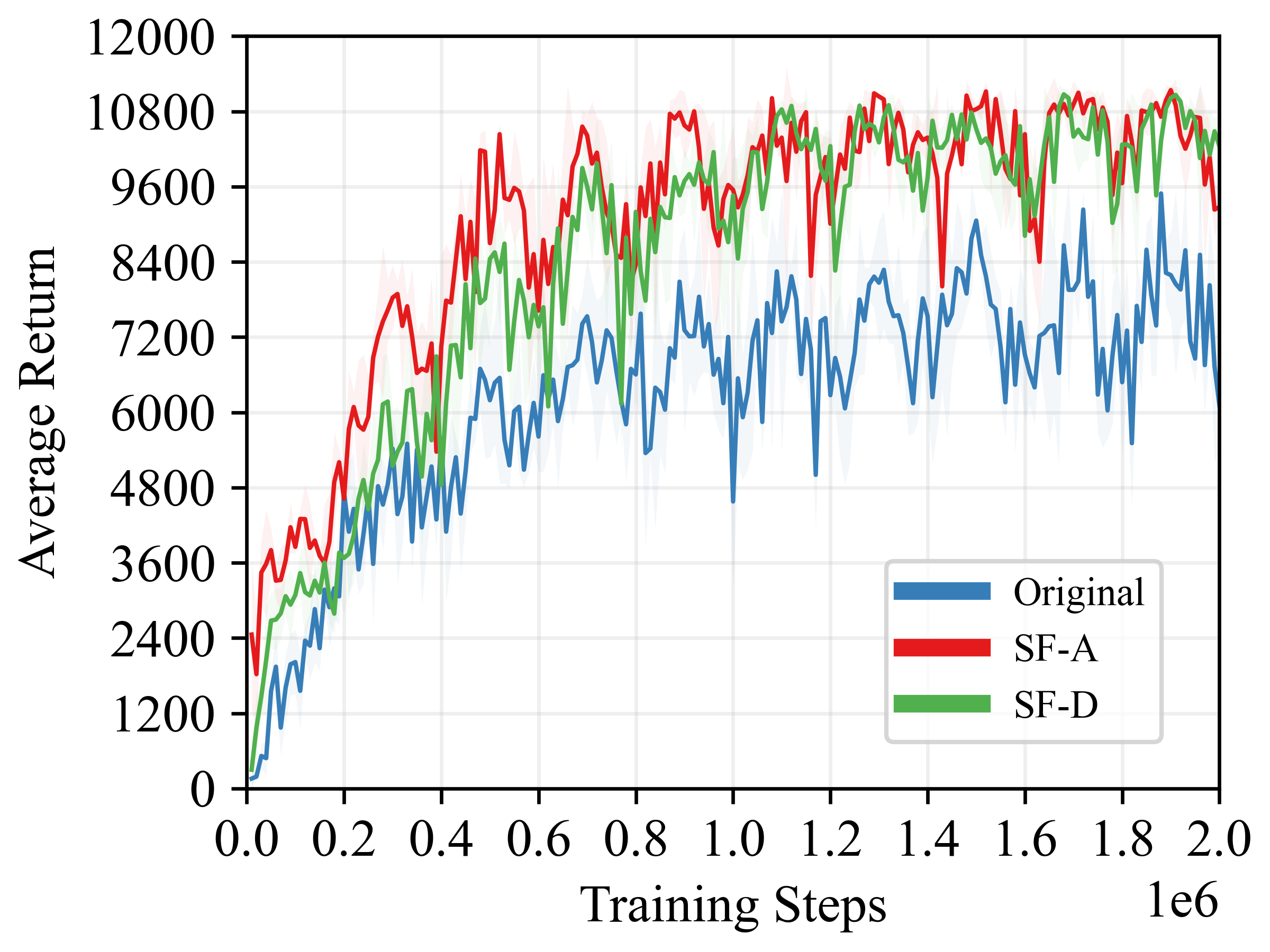} 
    \end{minipage}} 
    \\
    \subfloat[IQL-Hopper]{ 
    \begin{minipage}[b]{0.5\textwidth}
        \centering  
        \includegraphics[width=6cm, height=4.2cm]{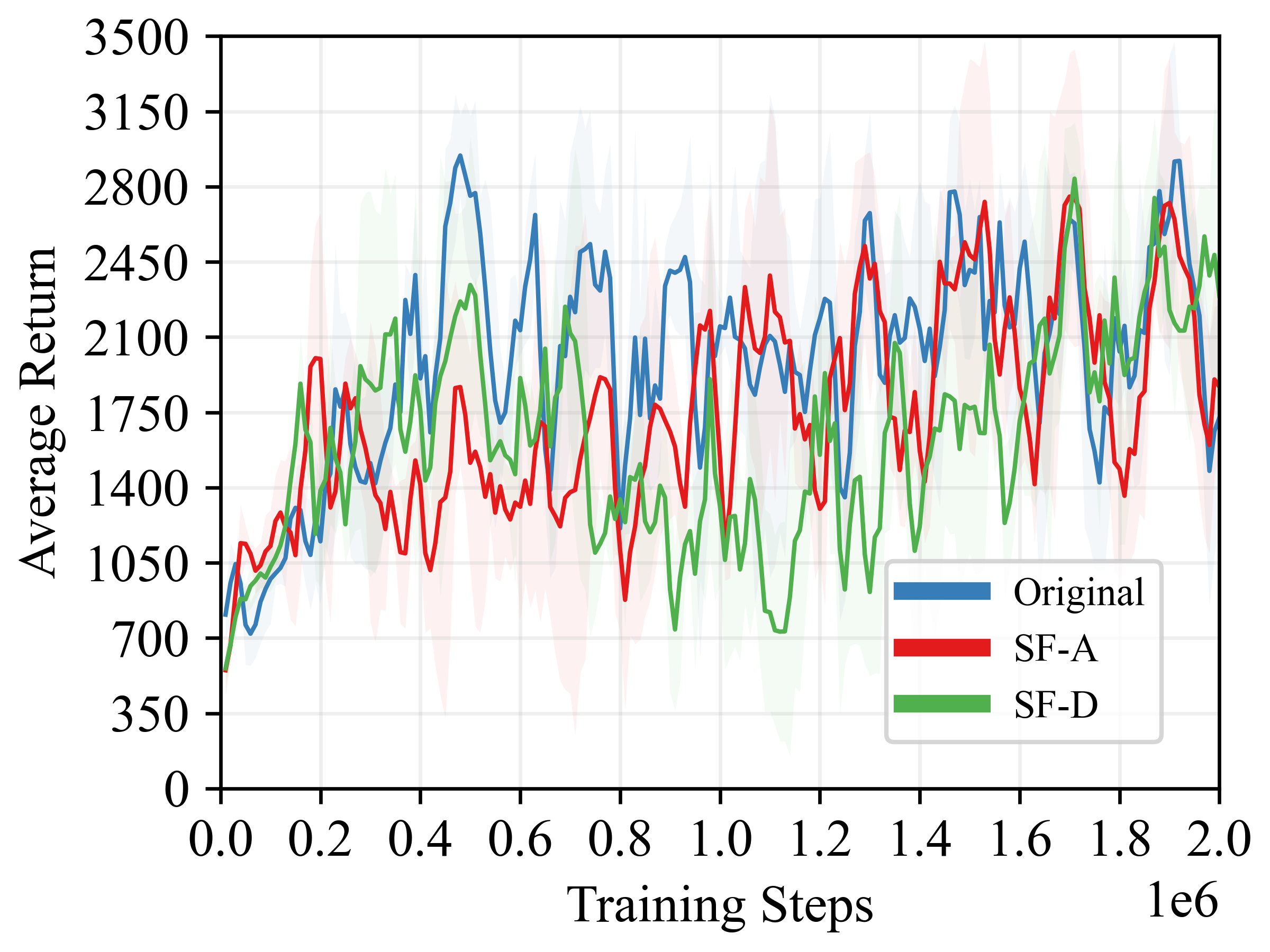} 
    \end{minipage}} 
       \subfloat[IQL-Walker2d]{ 
    \begin{minipage}[b]{0.5\textwidth}
        \centering  
        \includegraphics[width=6cm, height=4.2cm]{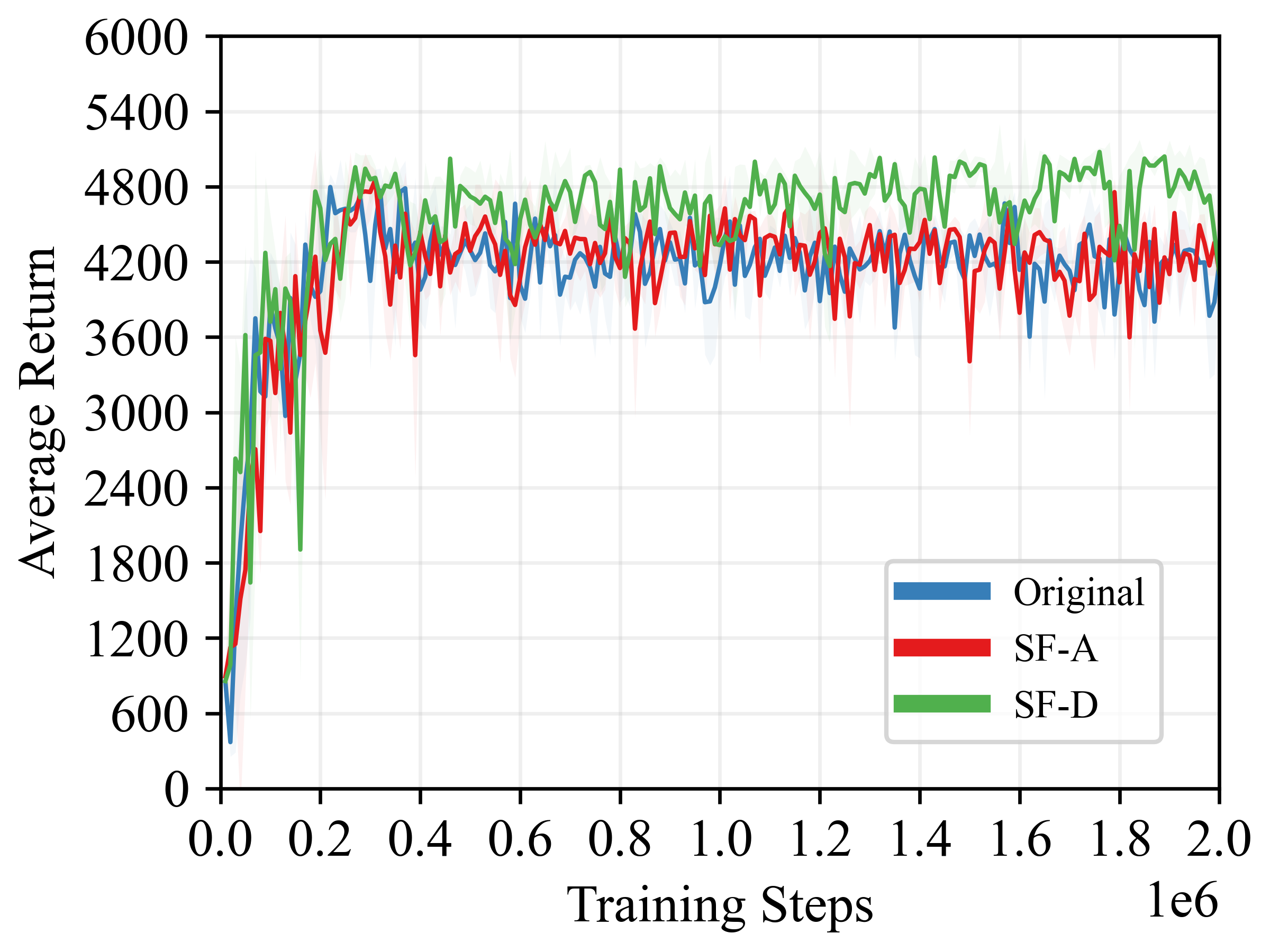} 
    \end{minipage}} 
    \caption{Experimental results of IQL.} 
    \label{Figure: Result IQL} 
\end{figure}

\begin{figure}[htbp] 
    \centering
    \subfloat[TD3+BC-Ant]{ 
    \begin{minipage}[b]{0.5\textwidth}
        \centering  
        \includegraphics[width=6cm, height=4.2cm]{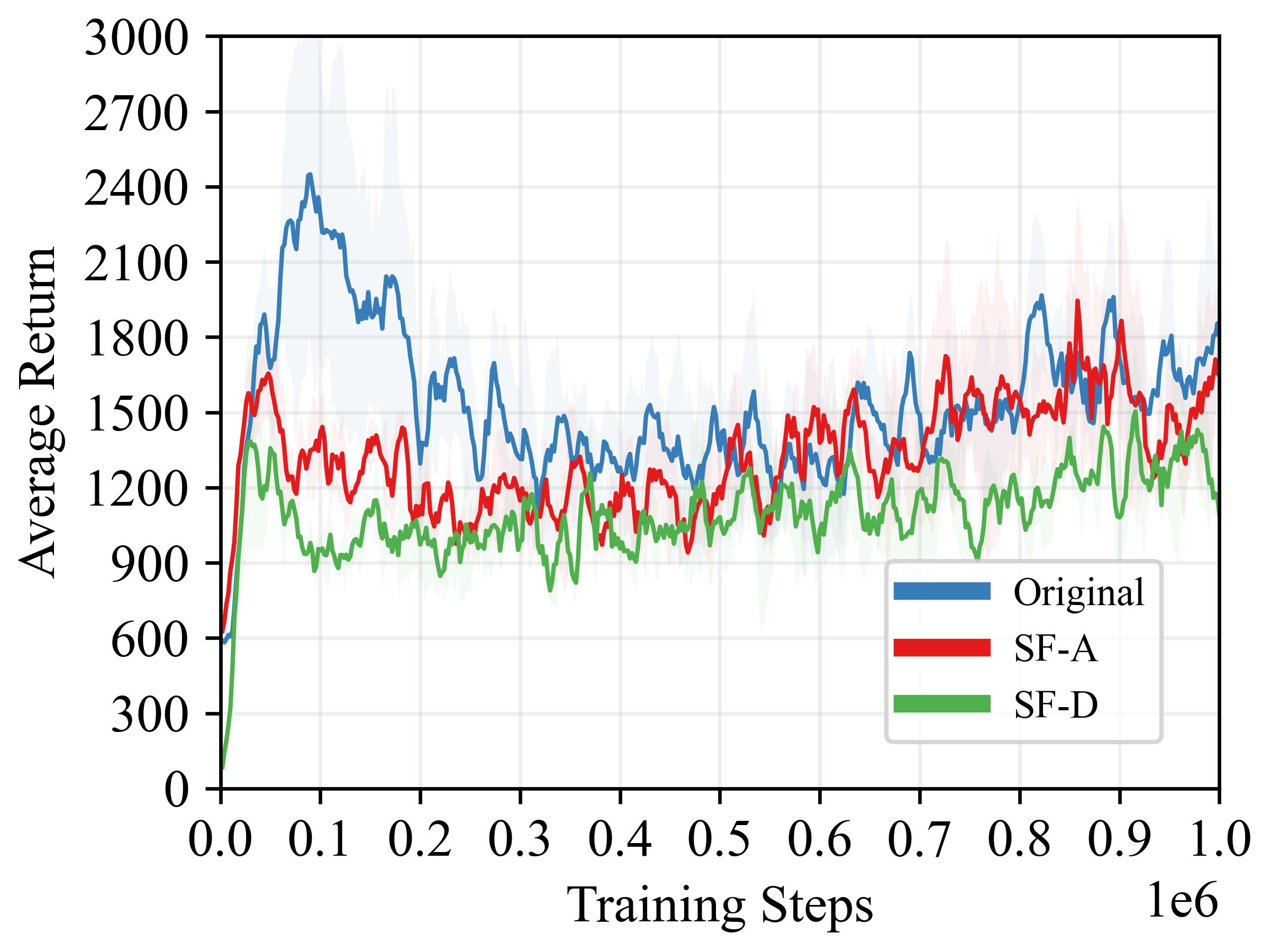} 
    \end{minipage}} 
    \subfloat[TD3+BC-HalfCheetah]{ 
    \begin{minipage}[b]{0.5\textwidth}
        \centering  
        \includegraphics[width=6cm, height=4.2cm]{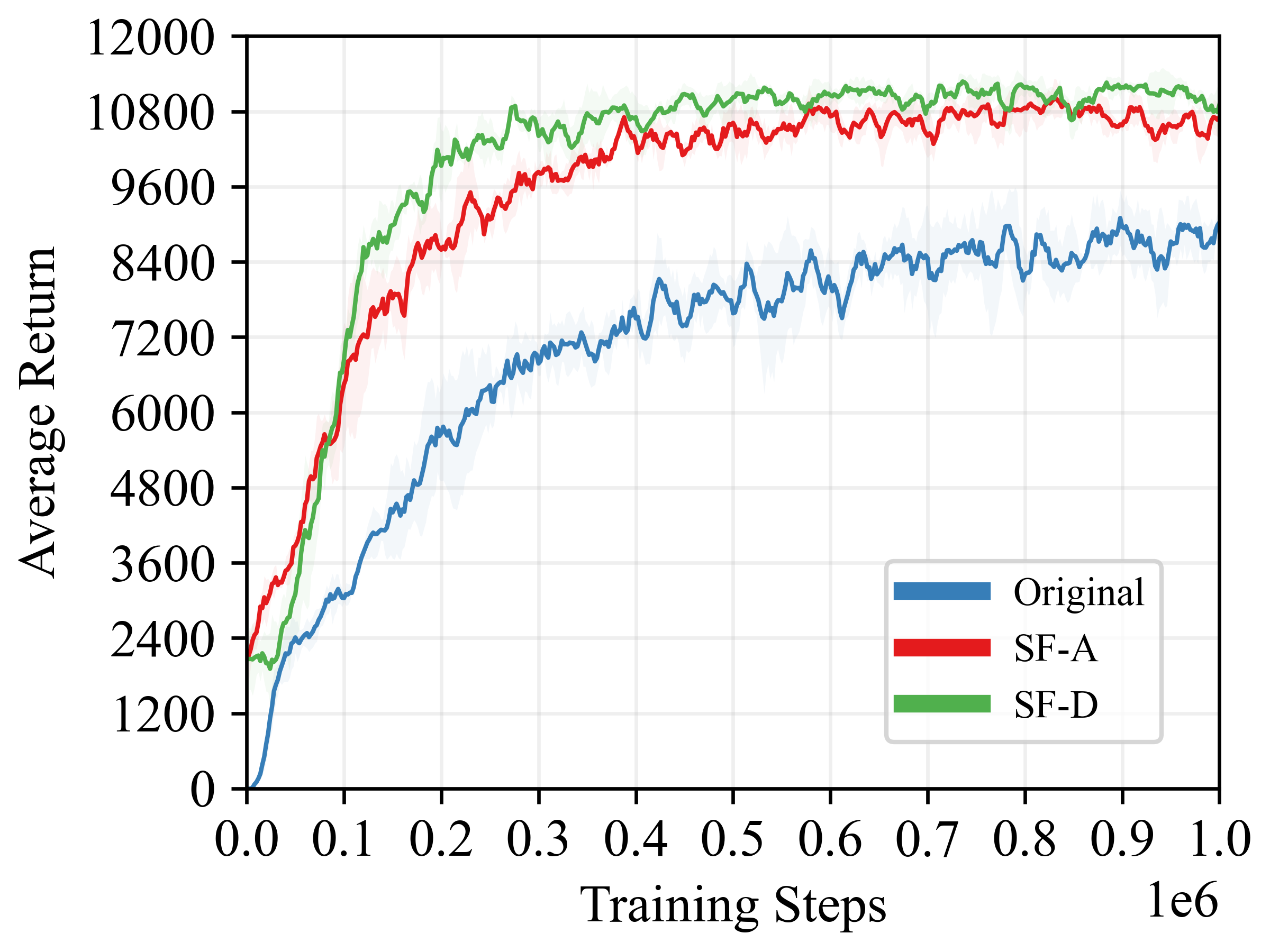} 
    \end{minipage}} 
    \\
    \subfloat[TD3+BC-Hopper]{ 
    \begin{minipage}[b]{0.5\textwidth}
        \centering  
        \includegraphics[width=6cm, height=4.2cm]{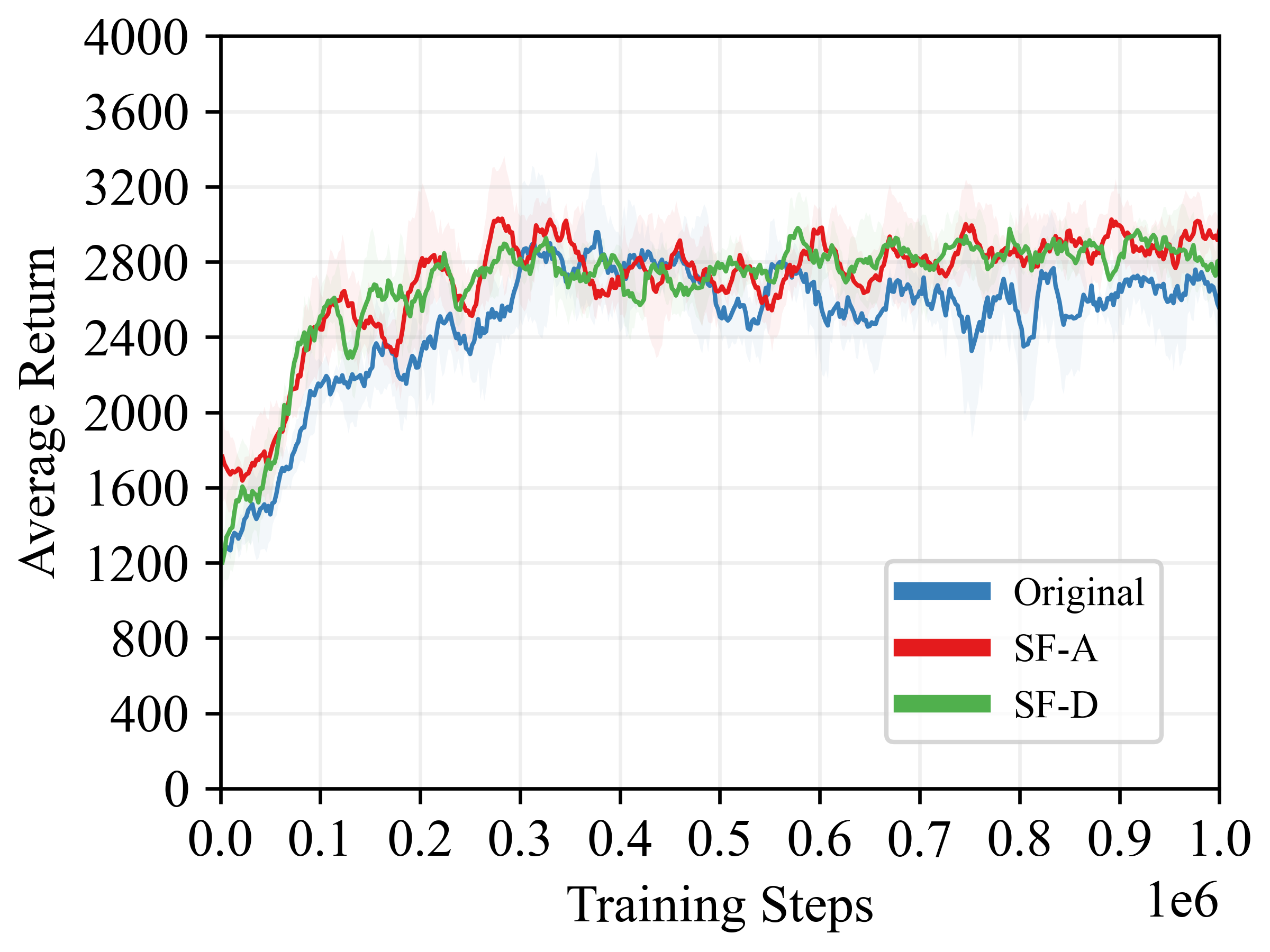} 
    \end{minipage}} 
       \subfloat[TD3+BC-Walker2d]{ 
    \begin{minipage}[b]{0.5\textwidth}
        \centering  
        \includegraphics[width=6cm, height=4.2cm]{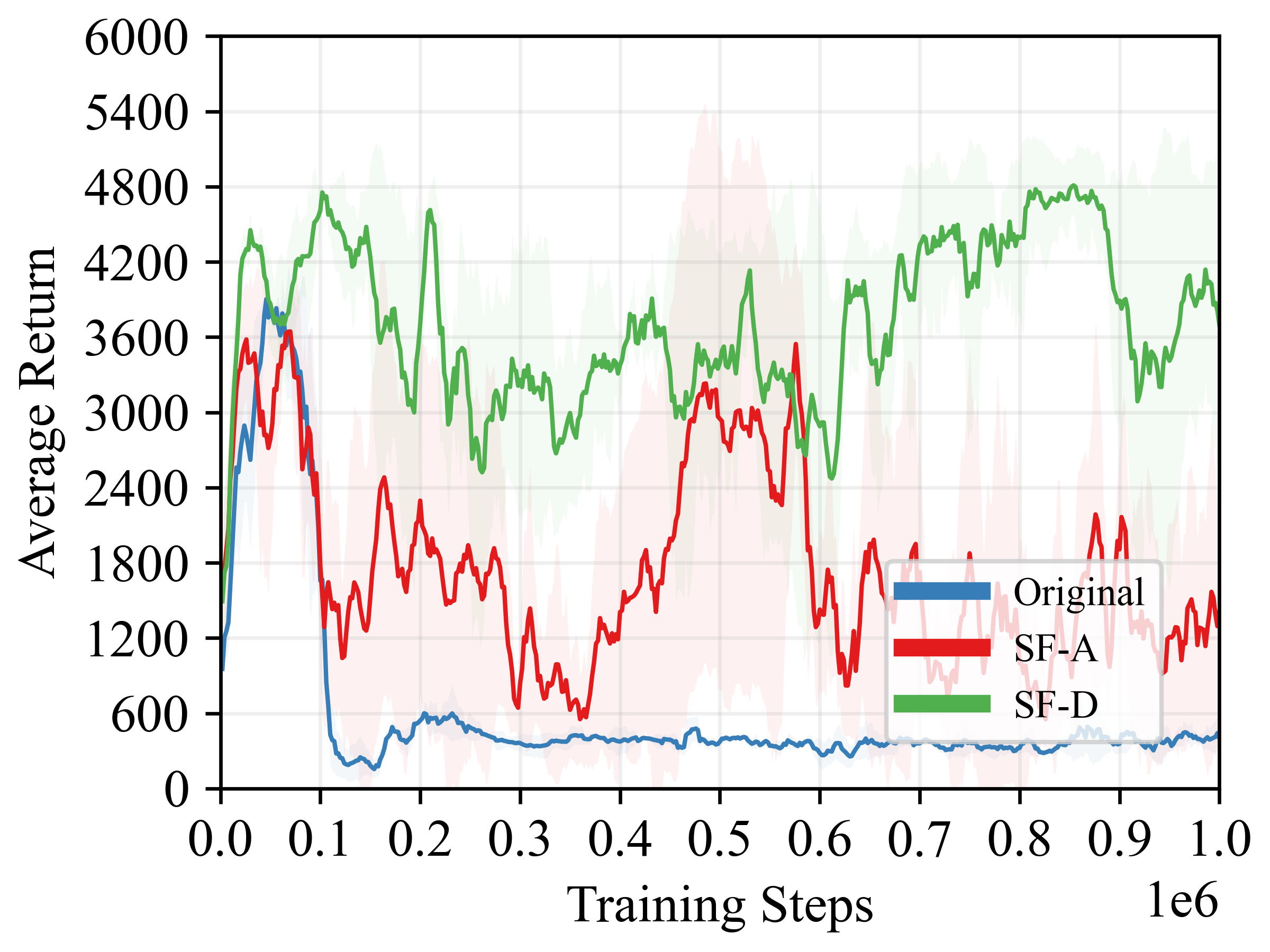} 
    \end{minipage}} 
    \caption{Experimental results of TD3+BC.} 
    \label{Figure: Result TD3BC} 
\end{figure}

\begin{table}[htbp]
    \centering
    \caption{Average returns}
    \label{tab: All}
    \begin{tabular}{c|c|c|c|c}
        \hline
        Environment          & Ant-v2           & HalfCheetah-v2            & Hopper-v2             & Walker2d-v2       \\
        \hline
        BEAR                & \textbf{3855.21}   & 656.98                   & 177.46                &1755.40   \\
        BEAR-SF-A(Ours)         & 3506.31             & 5576.45                & \textbf{233.73}        & 3106.24  \\
        BEAR-SF-D(Ours)        & 2898.11            & \textbf{7845.49}         &108.99                & \textbf{3355.45}       \\
        \hline
        IQL                 &4591.18            & 7473.01                   &2216.86                &4163.61           \\
        IQL-SF-A(Ours)          &\textbf{4695.44}    & 10154.54                 &2100.63                &4277.98          \\
        IQL-SF-D(Ours)         &4479.97            & \textbf{10539.98}         &\textbf{2296.91}       &\textbf{4704.72}         \\
        \hline
        TD3+BC(Ours)              &\textbf{1766.68}     & 8768.97                   &2648.98               &401.46          \\
        TD3+BC-SF-A(Ours)       &1615.81            & 10565.63                &\textbf{2936.52}                &1354.85           \\
        TD3+BC-SF-D(Ours)      &1243.54             & \textbf{10873.51}           &2766.64                 &\textbf{3934.22}        \\
        \hline
    \end{tabular}
\end{table}

\subsection{Results and Analysis}
The experimental results of BEAR, IQL and TD3+BC across four benchmark tasks are summarized in Table \ref{tab: All} and visualized in Figs. \ref{Figure: Result BEAR}-\ref{Figure: Result TD3BC}. Our proposed sample filtering variants are denoted as BEAR-SF-A/IQL-SF-A/TD3+BC-SF-A when using average reward criteria, and BEAR-SF-D/IQL-SF-D/TD3+BC-SF-D when employing average discounted reward criteria.

To ensure fair comparison of training efficiency, we maintain identical epoch counts between baseline methods and their filtered counterparts. Notably, the sample filtering mechanism inherently reduces dataset size, thereby decreasing the total training steps required while potentially improving data quality.

Table \ref{tab: All} and Figs. \ref{Figure: Result BEAR}-\ref{Figure: Result TD3BC} reveal three key observations: First, All three SF methods variants consistently outperform their original counterparts in three out of four tasks, regardless of filtering criteria. Second, while average reward and discounted reward criteria yield comparable returns in most environments, the latter demonstrates significant advantages in specific combinations, particularly in the HalfCheetah environment with BEAR and the Walker2d environment with TD3+BC. Third, in cases where the baseline algorithm demonstrates critically poor performance, such as the Hopper environment with BEAR, the application of SF methods fails to achieve statistically significant improvements in expected return.

When considering the visual aspect or analyzing the images in Figs. \ref{Figure: Result BEAR}-\ref{Figure: Result TD3BC}, it is noteworthy that despite being trained for fewer steps, this method achieves higher average returns. In the experiment, some of the curves of the proposed method display a quicker improvement and reach their peaks earlier than the originals. This can be attributed to the dataset filtering method, which eliminates poorer-quality data, enabling the model to learn better strategies from the outset. Additionally, the method shows smoother curves compared to the original method in specific environments, indicating reduced fluctuations in performance.

The empirical evidence collectively demonstrates that sample filtering effectively addresses two key challenges in policy-constrained offline RL: (1) mitigating performance degradation from low-quality samples through trajectory-level curation. (2) enabling faster policy improvement by focusing learning on high-return regions. This approach shows particular promise for real-world applications where datasets often contain heterogeneous quality trajectories, suggesting practical value in pre-processing pipelines for offline RL systems.

\section{Conclusions and Future Work}
\label{Section: Conclusions}
The paper proposed a simple yet effective sample filtering method designed to enhance both sample efficiency and the final performance of policy constraint offline RL algorithms. Experimental results on various offline RL algorithms and benchmark tasks proved that the proposed method outperforms baselines. However, it is essential to note that the method shows limitations as it primarily performs well with policy constraint offline RL algorithms, while its efficacy may be reduced when applied to other types of offline RL algorithms. For future work, we aim to employ more criteria for the sample filtering method and employ the proposed method to address real-world industry problems, further exploring its practical applications.

\section*{Acknowledgements}
This work was supported by the National Natural Science Foundation of China (Grant No. 62503100), China Postdoctoral Science Foundation (Grant No. 2025M771708), and the Shenzhen Basic Research Program (Grant No. JCYJ20220818102415033,  KJZD2023092311422045).

\appendix

\section{Implementation Details}
\label{Section: Implement details}

The implementation employs Python 3.8.0 with PyTorch 2.0.0 \cite{Ansel_PyTorch_2_Faster_2024} as the primary deep learning framework, supported by CUDA 11.7 \cite{10.5555/1891996} for GPU acceleration. We utilize D4RL 1.1 benchmark datasets with Mujoco-py 2.1.2.14 \cite{todorov2012mujoco} and Gym 0.21.0 \cite{1606.01540} for environment interactions, along with D3rlpy 1.1.1 \cite{d3rlpy} for offline reinforcement learning implementations. All experiments were conducted on a workstation equipped with an NVIDIA RTX 3090 GPU and Intel Xeon Gold 6240R CPU. For algorithm configurations, the parameters of the three algorithms used are shown in Tables \ref{tab: Hyperparameter Table of BEAR}-\ref{tab: Hyperparameter Table of TD3BC} respectively.

\begin{table}[htbp]
    \centering
    \caption{Hyperparameter table of BEAR}
    \label{tab: Hyperparameter Table of BEAR}
    \begin{tabular}{c|c|c}
        \hline
                       & Hyperparameter          & Value     \\
        \hline
                            &Optimizer            &Adam\cite{kingma2015adam}        \\
                             &Critic learning rate  &0.0001     \\
                            &Actor learning rate   &0.0003     \\
         BEAR Hyperparameters &Imitator learning rate   &0.0003     \\
                            &Mini-batch size       &256        \\
                            &Discount factor       &0.99       \\
                            &Target update rate    &0.005      \\
                            &MMD threshold         &0.05       \\
        \hline   
                        &Critic hidden dim     &256        \\
                            &Critic hidden layers  &2          \\
        Network Architecture  &Critic activation     &ReLU       \\
                            &Actor hidden dim      &256        \\
                            &Actor hidden layers   &2          \\
                            &Actor activation      &ReLU       \\
        \hline
    \end{tabular}
\end{table}

\begin{table}[htbp]
    \centering
    \caption{Hyperparameter table of IQL}
    \label{tab: Hyperparameter Table of IQL}
    \begin{tabular}{c|c|c}
        \hline
                       & Hyperparameter          & Value     \\
        \hline
                            &Optimizer            &Adam        \\
         &Q learning rate      &0.0003     \\
                            &V learning rate      &0.0003     \\
        IQL Hyperparameters                    &Mini-batch size      &256        \\
                            &Discount factor      &0.99       \\
                            &Temperature           &3.0        \\
                            &Expectile value      &0.7        \\
                            &Target update rate   &0.005      \\
        \hline   
        &Q hidden dim         &256        \\
                            &Q hidden layers      &2          \\
        Network Architecture                    &Q activation         &ReLU       \\
                            &V hidden dim         &256        \\
                            &V hidden layers      &2          \\
                            &V activation         &ReLU       \\
        \hline
    \end{tabular}
\end{table}

\begin{table}[htbp]
    \centering
    \caption{Hyperparameter table of TD3BC}
    \label{tab: Hyperparameter Table of TD3BC}
    \begin{tabular}{c|c|c}
        \hline
                       & Hyperparameter          & Value     \\
        \hline
                            &Optimizer            &Adam                        \\
         &Critic learning rate  &0.003                    \\
                            &Actor learning rate   &0.003                    \\
        TD3 Hyperparameters                    &Mini-batch size       &256                           \\
                            &Discount factor       &0.99               \\
                            &Target update rate     &0.005           \\
                            &Policy noise           &0.2                   \\
                            &Policy noise clipping    &(-0.5, 0.5)           \\
                            &Policy update frequency   & 2             \\
                            &Alpha                     &2.5\\
        \hline   
                &Critic hidden dim  &256\\
                                    &Critic hidden layers   &2\\
        Network Architecture                            &Critic activation function &ReLU\\
                                    &Actor hidden dim   &256\\
                                    &Actor hidden layers    &2\\
                                    &Actor activation function  &ReLU\\
        \hline
    \end{tabular}
\end{table}

\clearpage
\bibliographystyle{model1-num-names}
\bibliography{bibfile}

\end{sloppypar}
\end{document}